\documentclass{article}
\usepackage{amsmath,graphicx,mlspconf}
\usepackage{cite}
\usepackage{amsmath,amssymb,amsfonts}
\usepackage{algorithmic}
\usepackage{graphicx}
\usepackage{textcomp}
\usepackage{caption}
\usepackage{subcaption}
\usepackage{xcolor}
\usepackage{algorithm}
\usepackage{ragged2e}
\usepackage{siunitx}
\usepackage{url}
\usepackage{array}
\usepackage{float}
\usepackage{dblfloatfix}
\usepackage{multirow}
\usepackage{mathtools}
\usepackage{amssymb}

\DeclareMathOperator*{\argmax}{arg\,max} 
\DeclareMathOperator*{\argmin}{arg\,min} 
\newtheorem{lemma}{Lemma}
\toappear{2024 IEEE International Workshop on Machine Learning for Signal Processing, Sept.\ 22--25, 2024, London, UK}

\title{Negative Binomial Matrix Completion}
\name{%
   Yu Lu$^{\star}$%
   \qquad Kevin Bui$^{\dagger}$%
   \qquad Roummel F. Marcia$^{\star}$\thanks{This research is partially supported by NSF Grants IIS 1741490 and DMS 1840265.}%
}
\address{%
   $^{\star}$ Department of Applied Mathematics, University of California, Merced, California, USA \\%
   $^{\dagger}$ Department of Mathematics, University of California, Irvine, California, USA%
}

\begin{document}

\maketitle

\begin{abstract}
Matrix completion focuses on recovering missing or incomplete information in matrices. This problem arises in various applications, including image processing and network analysis. Previous research proposed Poisson matrix completion for count data with noise that follows a Poisson distribution, which assumes that the mean and variance are equal. Since overdispersed count data, whose variance is greater than the mean, is more likely to occur in realistic settings, we assume that the noise follows the negative binomial (NB) distribution, which can be  more general than the Poisson distribution. In this paper, we introduce NB matrix completion by proposing a nuclear-norm regularized model that can be solved by proximal gradient descent. In our experiments, we demonstrate that the NB model outperforms Poisson matrix completion in various noise and missing data settings on real data.
\end{abstract}

\begin{keywords}
Matrix completion, negative binomial distribution, data recovery, nonconvex optimization
\end{keywords}

\section{Introduction}
Matrix completion aims to recover the low-rank, ground-truth matrix $M \in \mathbb{R}^{m \times n}$ from an incomplete, noisy observation matrix $Y\in \mathbb{R}^{m \times n}$. Such situations occur widely in fields such as medical imaging \cite{hu2018generalized} and social network recovery \cite{mahindre2019sampling}.

Previous research has shown that nuclear norm minimization is an effective method for matrix completion\cite{cai2010singular, candes2012exact}, thereby serving as a foundation for many matrix completion algorithms\cite{bertsimas2020fast, cao2015poisson, candes2010matrix, tanner2013normalized}. Early works often developed matrix completion models and algorithms based on the noise-free assumption. Consequently, these models and algorithms were extended to more practical settings where the observations are noisy. For example, researchers redesigned them for cases when observations are corrupted with Gaussian noise or similar \cite{gunasekar2013noisy, keshavan2009matrix, keshavan2009low}. For count data, typically observed in photon-limited imaging \cite{harmany2010sparsity} and network traffic analysis \cite{bazerque2013inference}, the noise is commonly assumed to follow a Poisson distribution, as done in \cite{adhikari2015nonconvex, bui2023weighted, cao2015poisson, harmany2011spiral, le2007variational}. Adapted to this case, Poisson matrix completion was proposed in \cite{cao2015poisson}. However, the Poisson distribution requires the data's mean and variance to be equal, which is 
a strong assumption.  
Hence, we assume a more 
general 
distribution, the negative binomial (NB) distribution, for count data whose variance is greater than the mean\cite{anscombe1948transformation,gardner1995regression}. The NB noise assumption motivated further modifications to existing statistical models, such as regression \cite{allison2002, gardner1995regression, zhang2022elastic} and matrix factorization \cite{gouvert2020negative}. These models have been used in various applications, including traffic accident analysis \cite{poch1996negative}, genomics \cite{robinson2010edgeR}, and signal/image processing \cite{lu2023negative}.

Consisting of a sequence of independent and identically distributed Bernoulli trials, the $\text{NB}(r,p)$ distribution has the following probability mass function (p.m.f.):
\begin{align*}
    \mathbb{P}(y; r, p) = \binom{r+y-1}{y}(1-p)^y p^r,
\end{align*}
where $y \geq 0$ is the number of successful trials before the $r^{\text{th}}$ failure trial and $p \in [0,1]$ is the failure probability of each trial. The mean of the $\text{NB}(r,p)$ distribution is $\mu = r(1-p)/p$, so $p = r/(r+\mu)$. The p.m.f. of the $\text{NB}(r,p)$ can be shown to converge to the p.m.f. of the Poisson distribution with mean $\mu$ as $r \to \infty$ and $p \to 1$~\cite{degroot2012probability}. As a result, the Poisson distribution can be approximated by the NB distribution.

Let $M \in \mathbb{R}_+^{m \times n}$ be a matrix of non-negative entries, and let $\{ Y_{i,j} \}_{(i,j) \in \Omega}$ be a set of noisy observations of the entries of $M$  over an index set 
$\Omega \subseteq \{1, 2, \dotsc, m\} \times \{1, 2, \dotsc, n\}$. 
Under the NB assumption with fixed parameter $r$, the observed matrix $Y$ has the following entries:
\begin{gather}
\begin{aligned}\label{eq:NB model}
Y_{i,j} \sim \text{NB}\left(r, \frac{r}{r + M_{i, j}}\right), \ \ \ \ \text{for all }  (i, j) \in \Omega.
\end{aligned}
\end{gather}
Note that each entry $Y_{i,j}$ is independent of each other.
Our goal is to recover the low-rank, ground-truth matrix $M$ from the incomplete set $\{ Y_{i,j} \}_{(i,j) \in \Omega}$ corrupted with noise modeled as an NB distribution.

\section{Model Formulation and Algorithm}
In this section, we formulate a model to recover the ground-truth matrix $M$ from the partially observed  and NB-noise corrupted matrix $Y \in \mathbb{Z}_+^{m \times n}$ with entries from $\{ Y_{i,j} \}_{(i,j) \in \Omega}$. By \eqref{eq:NB model}, the probability of observing $Y$ given the ground-truth matrix $M$ is
\begin{gather}
\begin{aligned}\label{eq:NBObj}
    &\mathbb{P}(Y|M) =\\ &\prod_{(i,j) \in \Omega} \binom{r+Y_{i,j}-1}{Y_{i,j}} \left(\frac{M_{i, j}}{r + M_{i, j}}\right)^{Y_{i,j}} \left(\frac{r}{r + M_{i, j}}\right)^r.
\end{aligned}
\end{gather}
Then by Bayes' Theorem, the maximum a posteriori estimate of $M$ is obtained by
\begin{gather}
\begin{aligned}\label{eq:nb_map}
    \widehat{M} &= \argmax_X \mathbb{P}(X|Y) \\
    &= \argmax_X \frac{\mathbb{P}(Y|X) \mathbb{P}(X)}{\mathbb{P}(Y)} \\
    &=\argmin_X -\log \mathbb{P}(Y|X) - \log \mathbb{P}(X).
\end{aligned}
\end{gather}
Because 
\begin{align*}
    - \log \mathbb{P}(Y|X) = &\sum_{(i,j) \in \Omega} - \log((r+Y_{i,j}-1)!) + \log(Y_{i,j}!)\\
    &+ \log((r-1)!) - r \log(r) - Y_{i,j}\log(X_{i,j})\\
    &+ (r+Y_{i,j}) \log(r+X_{i,j}),
\end{align*}
\eqref{eq:nb_map} simplifies to
\begin{align}\label{eq:simple_nb_map}
    \widehat{M} = \argmin_X F(X)- \log \mathbb{P}(X),
\end{align}
where
\begin{align*}
F(X) \equiv \sum_{(i, j) \in \Omega} (r + Y_{i, j}) \log(r + X_{i, j}) - Y_{i,j} \log(X_{i, j}).
\end{align*}
The latter term $-\log \mathbb{P}(X)$ in \eqref{eq:simple_nb_map} is the log prior and can be interpreted as a regularizer. Since we desire $\widehat{M}$ to be low rank, we replace the log prior with nuclear norm regularization. Hence, our proposed model to recover the ground-truth matrix $M$ is
\begin{align} \label{eq:optimization problem}
\widehat{M} = \argmin_X \ F(X) + {\tau}  \|X\|_*,
\end{align}
where the nuclear norm $\|\cdot\|_*$ is the sum of the singular values of a matrix and $\tau > 0$ is a regularization parameter.

We solve \eqref{eq:optimization problem} by proximal gradient descent (see e.g., \cite{moreau1962fonctions,combettes2005signal}). In each iteration $k$, we linearize $F$ and solve the following optimization problem:
\begin{gather}\label{eq:linearize_opt}
\begin{aligned}
    X^{k+1} = \argmin_X &\;F(X^k) + \langle X - X^k, \nabla F(X^k) \rangle \\
&+ \frac{t_k}{2} \|X-X^k\|_F^2 + \tau \|X\|_*,
\end{aligned}
\end{gather}
where $t_k >0$ is the reciprocal of the step size (see \cite{cao2015poisson})
and $\|\cdot\|_F$ is the Frobenius norm. Note that each entry of $\nabla F(X^k)$ is given by
\begin{align*}
    (\nabla F(X^k))_{i,j} = \begin{cases} \displaystyle \frac{r + Y_{i,j}}{r+X^k_{i,j}} - \frac{Y_{i,j}}{X^k_{i,j}} &\text{ if } (i,j) \in \Omega,\\
    0 &\text{ if } (i,j) \not \in \Omega.
    \end{cases}
\end{align*} By rearranging \eqref{eq:linearize_opt}, we obtain
\begin{gather}\label{eq:rearrange}
\begin{aligned}
    X^{k+1} &= \argmin_X \|X\|_* + \frac{1}{2 \frac{\tau}{t_k}}\left \|X- X^k + \frac{1}{t_k}\nabla F(X^k) \right\|_F^2 \\
    &= \text{prox}_{\frac{\tau}{t_k} \|\cdot\|_*} \left(X^k - \frac{1}{t_k}\nabla F(X^k) \right), 
\end{aligned}
\end{gather}
where the proximal operator of $\|\cdot\|_*$ is defined by
\begin{align}\label{eq:proximal}
    \text{prox}_{\lambda \|\cdot\|_*} (Z) = \argmin_X \|X\|_* + \frac{1}{2\lambda} \|X-Z\|_F^2, 
\end{align}
for $\lambda >0.$ 
The exact solution to \eqref{eq:proximal} is provided by the following lemma:
\begin{lemma}[Theorem 2.1, \cite{cai2010singular}] \label{lemma:prox}
    Consider the singular value decomposition of $Z\in \mathbb{R}^{m \times n}$ of rank $l$ as follows:
    \begin{align*}
        Z = U\Sigma V^{\top},\; \Sigma = \text{diag}(\{\sigma_i\}_{i=1}^l),
    \end{align*}
    where $U \in \mathbb{R}^{m \times l}$ and $V \in \mathbb{R}^{n \times l}$ are matrices with orthonormal columns. For $\lambda > 0$, let
    \begin{align*}
        \mathcal{D}_{\lambda}(Z) \coloneqq U \mathcal{D}_{\lambda}(\Sigma)V^{\top}, \; \mathcal{D}_{\lambda}(\Sigma) = \text{diag}\left(\left\{\left(\sigma_i - \lambda\right)_+\right\}_{i=1}^l\right),
    \end{align*}
    where $(\sigma -\lambda)_+ = \max(0,\sigma-\lambda)$. Then
    \begin{align*}
        \mathcal{D}_{\lambda}(Z) = \text{prox}_{\lambda \|\cdot\|_*} (Z).
    \end{align*}
\end{lemma}
Therefore, by Lemma \ref{lemma:prox}, \eqref{eq:rearrange} simplifies to
\begin{align}
    X^{k+1} = \mathcal{D}_{\frac{\tau}{t_k}}(Z^k),
\end{align}
where $Z^k = X^k - \frac{1}{t_k} \nabla F(X^k)$. Because $\frac{1}{t_k}$ is the step size in the gradient descent step in \eqref{eq:rearrange}, we increase $t_k$ by a factor $\eta > 0$ whenever $F(X^{k+1}) > F(X^k)$ to ensure that $\{F(X^k)\}_{k=0}^{\infty}$ is monotonically decreasing. The overall algorithm is summarized in Algorithm \ref{alg:alg}.

\begin{algorithm}[t!!!]
\caption{Proximal Gradient Descent for NB Matrix Completion}
\label{alg:alg}
\scriptsize
\begin{algorithmic}[1]
\REQUIRE Noisy incomplete matrix $Y$, upper bound $\alpha>0$, lower bound $\beta>0$, regularization parameter $\tau$, penalty multiplier $\eta > 1$,  index set $\Omega$, the maximum iteration $K$, the stopping criteria $tol$ 
    \STATE Initialize $X^0_{i,j} = Y_{i,j}\ \text{if } (i,j) \in \Omega$\ \text{else } $X^0_{i,j} = \frac{\alpha + \beta}{2}$
    \STATE $k=0$ 
    \WHILE{$k \leq K$ and $tol < err$} 
    \STATE $Z^k = X^k - \frac{1}{t_k} \nabla F(X^k)$
    \STATE $X^{k+1} = \mathcal{D}_{\frac{\tau}{t_k}}(Z^k)$
    \IF{$F(X^{k+1}) > F^k(X^{k})$}
        \STATE $t_k = \eta t_k$
        \STATE \text{Go back to 4}
    \ENDIF
    \STATE $err = F(X^{k+1}) - F(X^{k})$
    \STATE $k = k + 1$
    \ENDWHILE
    \RETURN Recovered matrix $\widehat{M}=X^{k}$. \\
\end{algorithmic}
\end{algorithm}

\section{Experiment}

We conduct experiments on three different datasets corrupted with either Poisson noise or NB noise with dispersion parameters $r=10$ and $r=25$. For each noise level and dataset, we consider three different percentages $q$ of known entries in the matrix, with $q = 25\%, 50\%$, and $75\%$. In each experiment, we compare our proposed NB model \eqref{eq:optimization problem} with its Poisson counterpart \cite{cao2015poisson}. Each experiment was conducted 20 times with different noisy realizations, and the average of these trials is presented. For numerical evaluation, we use peak signal-to-noise ratio (PSNR) and normalized root-mean-square error (NRMSE), given as follows:
\begin{align*}
\text{PSNR} &= 10 \cdot \log_{10} \left(\frac{M_{\text{max}}^2\cdot mn }{
S
}
\right)
\end{align*}
and
\begin{align*}
    \text{NRMSE} &= \frac{1}{M_{\text{max}}-M_{\text{min}}} \sqrt{\frac{S}{mn}
    },
\end{align*}
where $S = \sum_{i=1}^m \sum_{j=1}^{n} \left(\widehat{M}_{i,j} - M_{i,j}\right)^2
$ and $M_{\text{min}}$ and $M_{\text{max}}$ are the minimum and maximum entry values of the ground-truth matrix $M$, respectively.  

The parameters for Algorithm \ref{alg:alg} are set as follows: maximum iteration $K=2000$, the stopping criterion $tol=1e-6$, and the penalty multiplier $\eta=1.1$. The bounds $\alpha$ and $\beta$ are chosen based on the maximum and minimum values of the entries, respectively. The regularization parameter $\tau$ is searched within the set $\{10, 20, \ldots, 310\}$.

For the NB models, several methods exist for estimating the parameter $r$, such as the method-of-moments \cite{clark1989estimation}, the maximum quasi-likelihood methods \cite{piegorsch1990maximum}, and cross-validation techniques. However, to eliminate potential biases or inaccuracies inherent in the parameter estimation process, our experiments intentionally use the exact value of the parameter $r$ as a prior. In the cases of Poisson noise, we set $r=1000$ to have our NB model approximate the Poisson model.

One point that needs to be highlighted is that NB noise with $r=10$ is not always noisier than $r=25$. By \eqref{eq:NB model}, the noise level depends on both the dispersion parameter $r$ and the original matrix entries. For example, if the dispersion parameter $r$ is close to the original data values, the noise level would be minimal at these entries. Conversely, when the dispersion parameter $r$ is distant from the original data values, the corresponding entries can be extremely noisy. 

\subsection{Experiment 1: Bike-Sharing Count Data}\label{subsec:exp1}
\begin{table}[!h]
\scriptsize
\caption{Experiment 1 results. PSNRs and NRMSEs of the Poisson model and our proposed NB model on the Capital bike-sharing data with three different known-data level and three different noise level.  \textbf{Bold} indicates best value.}
\label{table:exp1}
\centering
\begin{tabular}{|c|c|
p{1.52cm}|
p{1.45cm}|
p{1.45cm}|
}
\cline{1-5}
\multirow{2}{*}{\ \  Mask}                 
& \multirow{2}{*}{\ \ Model }
& \multicolumn{3}{c|}{PSNR $(\uparrow)$ / NRMSE $(\downarrow)$} \\ \cline{3-5}
&
& \centering $r = 10$
& \centering  $r= 25$
& \quad \ Poisson
\\ \cline{1-5}    
\multirow{2}{*}{$q = 25\%$}
& Poisson
& 16.47 / 15.13\%
& 20.37 / 9.59\%
& \textbf{22.29} / \textbf{7.71\%} 
\\ 
\cline{2-2} 
& NB
& \textbf{18.05} / \textbf{12.52\%}
& \textbf{21.06} / \textbf{8.87\%} 
& 21.92 /  8.04\%
\\ 
\cline{1-5}
\multirow{2}{*}{$q = 50\%$}
& Poisson 
& 17.72 / 13.03\%
& 21.69 / 8.26\%  
& \textbf{27.12} / \textbf{4.41\%}
\\
\cline{2-2}
& NB
& \textbf{18.41} / \textbf{12.08\%}
& \textbf{22.58} / \textbf{7.63\%}
& 26.75 / 4.60\%
\\
\cline{1-5}
\multirow{2}{*}{$q = 75\%$}
& Poisson 
& 19.12 / 11.09\%
& 22.74 / 7.30\%
& \textbf{30.12} / \textbf{3.12\%}
\\
\cline{2-2}
& NB 
& \textbf{21.60} / \ \ \textbf{8.32\%} 
& \textbf{24.63} / \textbf{5.88\%}
& 29.65 / 3.30\% 
\\
\cline{1-5}
\end{tabular}
\end{table}
We examine the bike-sharing data aggregated on an hourly basis on Saturdays between 2011 and 2012 within the Capital Bike Share system\footnote{Accessed at \url{https://doi.org/10.24432/C5W894}}, resulting in an $24 \times 105$ matrix. The average count is 194 and the median count is 163, so NB noise with $r=10$ will be stronger than with $r=25$. The results of the models applied to this dataset is provided in Table \ref{table:exp1}.

Across all three masking levels, the NB model demonstrates more accurate results than the Poisson model when the data is corrupted by NB noise with $r=10,25$. However, for Poisson noise, the Poisson model slightly outperforms the NB model as expected: the difference in performance is only at most 0.50 for PSNR and 0.33\% for NRMSE. The most significant difference between the proposed NB model and the Poisson model occurs at the highest noise level, $r=10$, indicating that the NB model yields more accurate results in complex noisy situations than the Poisson model. As the level of missing data increases, we observe a decrease in accuracy for both the Poisson and the proposed NB model.


\subsection{Experiment 2: Vehicle Count Data}
\begin{figure}[!h]
\centering
\includegraphics[trim={.28cm 0 0 0},clip,scale=0.635]{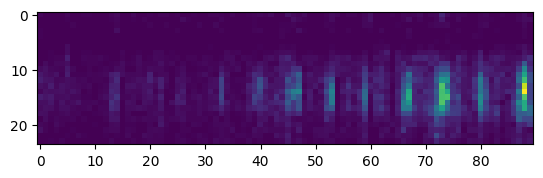}
\caption{Hourly inbound traffic data at Robert F. Kennedy Bridge Queens/Bronx Plaza from January 2021 to March 2021.}
\label{fig:original}
\end{figure}

We examine the hourly inbound traffic data at Robert F.\ Kennedy Bridge Queens/Bronx Plaza between January 2021 and March 2021\footnote{Accessed at \url{https://data.ny.gov/Transportation/Hourly-Traffic-on-Metropolitan-Transportation-Auth/qzve-kjga/about_data}}. The traffic data records the number of vehicles that passed by without the E-ZPass tag or with dysfunctional tags. Note that the the mean and median of the dataset are 12 and 4, respectively. Like in 
Experiment 1,
the data is formatted as a $24 \times 90$ matrix, shown in Fig.\  \ref{fig:original}.

\begin{figure}[!b]
     \centering
     \begin{subfigure}[b]{0.5\textwidth}
         \centering
         \includegraphics[trim={.28cm 0 0 0},clip,scale=0.635]{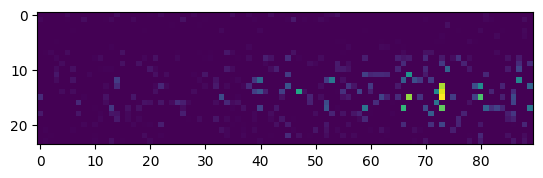}
         \caption{Data with Poisson noise and $25\%$  known entries}
         \label{fig:poisson_noisy}
     \end{subfigure}\\
     
          \begin{subfigure}[b]{0.5\textwidth}
         \centering
         \includegraphics[trim={.28cm 0 0 0},clip,scale=0.635]{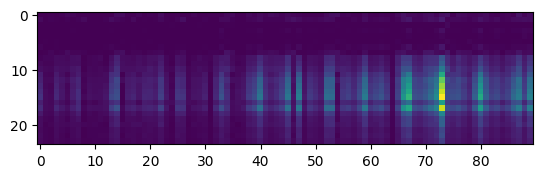}
         \caption{Poisson Reconstruction}
         \label{fig:poisson_result1}
     \end{subfigure}\\
     
               \begin{subfigure}[b]{0.5\textwidth}
         \centering
         \includegraphics[trim={.28cm 0 0 0},clip,scale=0.635]{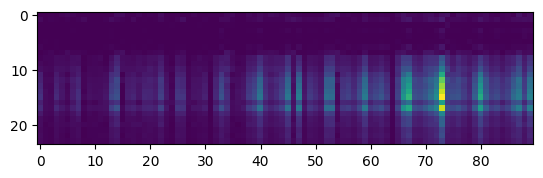}
         \caption{NB Reconstruction}
         \label{fig:nb_result1}
     \end{subfigure}
\caption{
Experiment 2 (Poisson) results. (a) Traffic data 
 (Fig.~\ref{fig:original}) corrupted with Poisson noise and only 25\% of its entries are known.
(b) Poisson reconstruction with PSNR $= 22.99$ and RMSE $= 7.09\%$. (c) NB reconstruction with PSNR $= 22.97$ and RMSE $= 7.11\%$.}
\label{fig:25Poisson}
\end{figure}

\begin{figure}[!b]
     \centering
     \begin{subfigure}[b]{0.5\textwidth}
         \centering
         \includegraphics[trim={.28cm 0 0 0},clip,scale=0.635]{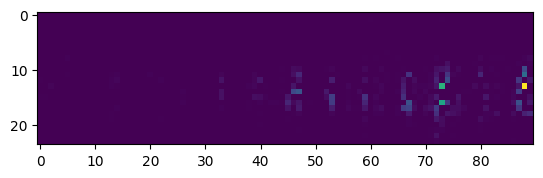}
         \caption{Data with NB ($r=10$) noise and $25\%$ known entries}
         \label{fig:nb_noisy}
     \end{subfigure}\\
     
          \begin{subfigure}[b]{0.5\textwidth}
         \centering
         \includegraphics[trim={.28cm 0 0 0},clip,scale=0.635]{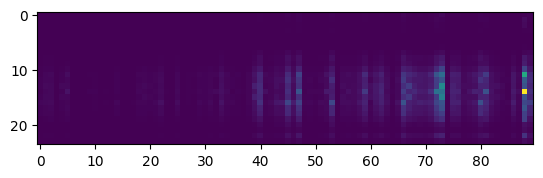}
         \caption{Poisson Reconstruction}
         \label{fig:poisson_result2}
     \end{subfigure}\\
     
               \begin{subfigure}[b]{0.5\textwidth}
         \centering
         \includegraphics[trim={.28cm 0 0 0},clip,scale=0.635]{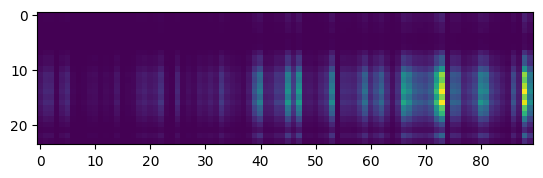}
         \caption{NB Reconstruction}
         \label{fig:nb_result2}
     \end{subfigure}
\caption{
Experiment 2 (NB) results. (a) Traffic data 
 (Fig.\  \ref{fig:original}) corrupted with NB ($r=10$) noise and only 25\% of its entries are known. (b) Poisson reconstruction with  PSNR $= 21.44$ and RMSE $= 8.47\%$. (c) NB reconstruction with  PSNR $= 23.97$ and RMSE $= 6.37\%$.}
\label{fig:25r10}
\end{figure}

\begin{table}[!ht]
\scriptsize
\caption{Experiment 2 results.  PSNRs and NRMSEs of the Poisson model and our proposed NB model on the hourly inbound traffic data at Robert F. Kennedy Bridge Queens/Bronx Plaza with three different known-data level and three different noise level.  \textbf{Bold} indicates best value.}
\label{table:exp2}
\centering
\begin{tabular}{|l|l|
>{\arraybackslash}p{1.25cm}|
>{\arraybackslash}p{1.25cm}|
>{\arraybackslash}p{1.25cm}|
}
\cline{1-5}
\multirow{2}{*}{\ \  Mask}                 
& \multirow{2}{*}{\ \ Model }
& \multicolumn{3}{c|}{PSNR $(\uparrow)$/NRMSE $(\downarrow)$} 
 \\ \cline{3-5}
&
& $r = 10$
& $r= 25$
& Poisson
\\ \cline{1-5}    
\multirow{2}{*}{$q = 25\%$}
& Poisson
& 21.44/8.47\%
& 21.22/8.70\%
& \textbf{22.99}/\textbf{7.09\%}      
\\ 
\cline{2-2} 
& NB
& \textbf{23.97}/\textbf{6.37\%}
& \textbf{22.44}/\textbf{7.59\%}
& 22.97/7.11\%
\\ 
\cline{1-5}
\multirow{2}{*}{$q = 50\%$}
& Poisson 
& 27.72/4.14\%
& 22.17/7.80\%
& \textbf{25.62}/\textbf{5.24\%}  
\\
\cline{2-2}
& NB
& \textbf{28.36}/\textbf{3.82\%}
& \textbf{23.75}/\textbf{6.50\%}
& 25.58/5.27\%
\\
\cline{1-5}
\multirow{2}{*}{$q = 75\%$}
& Poisson 
& 28.69/3.68\%
& 23.59/6.62\%
& \textbf{27.16}/\textbf{4.39\%} 
\\
\cline{2-2}
& NB 
& \textbf{29.79}/\textbf{3.24\%}
& \textbf{26.47}/\textbf{4.80\%}
& 27.10/4.42\% 
\\
\cline{1-5}
\end{tabular}
\end{table}

Table \ref{table:exp2} presents the PSNRs and NRMSEs of the Poisson and NB models.
As expected, we observe that the NB model is more accurate than the Poisson model for the NB noise cases, while the Poisson model achieves slightly better results for the Poisson noise case. The difference in performance for the Poisson noise case is less than 0.1 in PSNR and less than 0.04\% in NRMSE. Across all noise levels, the results for $q=75\%$ are more accurate than those for $q=25\%$ because the more available information by the observed entries improves the inference of the missing entries.  Since the dataset values are closer to 10 than to 25, $r=25$ results in a noisier matrix. Consequently, the results for $r=10$ are more accurate than those for $r=25$.

For the case when $q=25\%$ of the data entries are known, Fig.\ \ref{fig:25Poisson} visualizes the reconstruction results of the data matrix corrupted with Poisson noise, while Fig.\ \ref{fig:25r10} visualizes the reconstruction results of the data matrix corrupted with NB $(r=10)$ noise.  In Fig.\  \ref{fig:25Poisson}, the reconstruction results between the Poisson and NB model appear very similar, as validated by their nearly equal PSNRs and RMSEs. On the other hand, in Fig.\ \ref{fig:25r10}, the difference between the two reconstruction results is apparent. The NB reconstruction significantly resembles the original data matrix more than the Poisson reconstruction. Moreover, the NB reconstruction recovers stronger signals in the traffic count data than the Poisson reconstruction around the 73th and 89th days of 2021. Such signals are not clear in the corrupted matrix shown in Fig.\ \ref{fig:nb_noisy}. Overall, our proposed NB model is a more general and robust model for recovering realistic count data.


\subsection{Experiment 3: Microscopy Image}

\begin{figure}[!h]
     \centering
     \begin{subfigure}[b]{0.225\textwidth}
         \centering
         \includegraphics[scale=0.35]{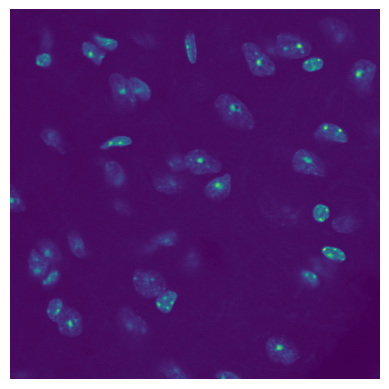}
         \caption{Low-Rank Image}
         \label{fig:lowrank}
     \end{subfigure}
     \begin{subfigure}[b]{0.225\textwidth}
         \centering
         \includegraphics[scale=0.35]{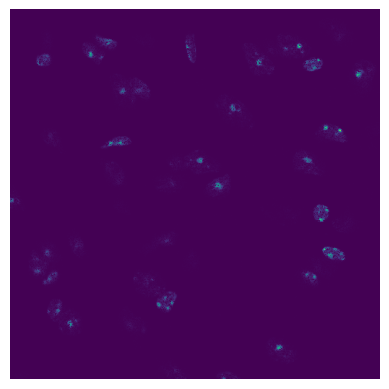}
         \caption{Noisy Image}
         \label{fig:noisyimage}
     \end{subfigure}\\
              \begin{subfigure}[b]{0.225\textwidth}
         \centering
         \includegraphics[scale=0.35]{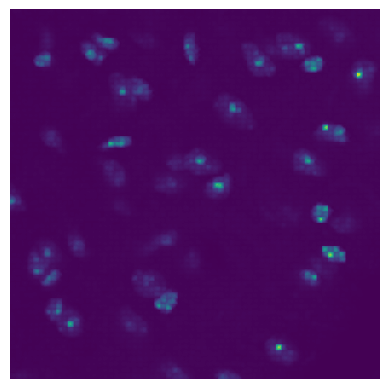}
         \caption{Poisson Reconstruction}
         \label{fig:poissonimage}
     \end{subfigure}
     \begin{subfigure}[b]{0.225\textwidth}
         \centering
         \includegraphics[scale=0.35]{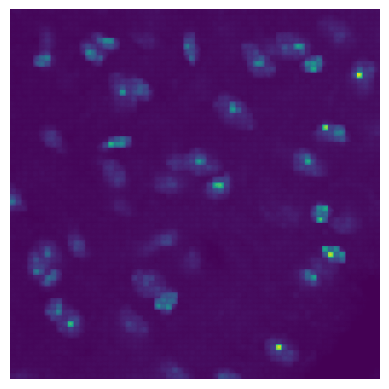}
         \caption{NB Reconstruction}
         \label{fig:nbimage}
     \end{subfigure}
 
        \caption{Results on the microscopy image of mouse brain tissues 
        from 
        \cite{zhang2019Poisson}. (a) Low-rank image of the original. (b) Incomplete, noisy image with NB noise $r=10$ NB noise level and $q=75\%$ data-known level. (c) Poisson reconstruction with PSNR $= 22.36$ and NRMSE $= 7.62\%$.  (d) NB model reconstruction with PSNR $= 29.08$ and RMSE $= 3.51\%$.}
        \label{fig:Exp3}
\end{figure}

We examine a $512 \times 512$ microscopy image of mouse brain tissues \cite{zhang2019Poisson}. The maximum pixel intensity value of this data is $242$, the minimum value is $2$, and the mean and median pixel intensity values are $14$ and $10$, respectively. Here, we break the image into $8 \times 8$ patches, vectorize them, and concatenate them into a $64 \times 4096$ matrix, and then approximate it by low-rank matrix with rank $l = 32$. This low-rank matrix is the ground-truth matrix $M$, shown in Fig.\ \ref{fig:lowrank}, for this experiment. Subsequently, we corrupt Fig.\  \ref{fig:lowrank} by NB noise with $r=10$ and observe $q=75\%$ of the entries. The Poisson and NB reconstructed images are shown in Fig.\  \ref{fig:poissonimage}-\ref{fig:nbimage}, where the Poisson reconstruction has $\text{PSNR} = 22.36$ and $\text{NRMSE}= 7.62\%$ while the NB reconstruction has $\text{PSNR} = 29.08$ and $\text{NRMSE}= 3.51\%$, demonstrating the improved performance of the proposed NB model over the Poisson model for this experiment.

\section{Conclusion}
We proposed NB matrix completion to accommodate overdispersed count data often encountered in real-world scenarios. To perform NB matrix completion, we introduced a model formulated using maximum a posteriori and nuclear norm regularization and solved it by proximal gradient descent. Through three experiments, we compared our proposed model with its Poisson counterpart. Our experiments demonstrated that the NB model significantly outperformed the Poisson model when the data is corrupted with NB noise. On the other hand, when the data is corrupted with Poisson noise, the NB model with large dispersion parameter $r$ maintained comparable performance as the Poisson model. In extreme cases of information loss, both the NB and Poisson models faced challenges in achieving high accuracy results.

\bibliographystyle{IEEEbib}
\bibliography{strings,refs}

\end{document}